# Revisiting Technical Bias Mitigation Strategies

Abdoul Jalil Djiberou Mahamadou, Artem A. Trotsyuk
Center for Biomedical Ethics, Stanford University School of Medicine, Stanford, California, USA
emails: abdjiber@stanford.edu, atrotsyuk@stanford.edu

**Abstract**

Efforts to mitigate bias and enhance fairness in the artificial intelligence (AI) community have predominantly focused on technical solutions. While numerous reviews have addressed bias in AI, this review uniquely focuses on the practical limitations of technical solutions in healthcare settings, providing a structured analysis across five key dimensions affecting their real-world implementation: who defines bias and fairness; which mitigation strategy to use and prioritize among dozens that are inconsistent and incompatible; when in the AI development stages the solutions are most effective; for which populations; and the context in which the solutions are designed. We illustrate each limitation with empirical studies focusing on healthcare and biomedical applications. Moreover, we discuss how value-sensitive AI, a framework derived from technology design, can engage stakeholders and ensure that their values are embodied in bias and fairness mitigation solutions. Finally, we discuss areas that require further investigation and provide practical recommendations to address the limitations covered in the study.

**Introduction**

Artificial intelligence (AI) is becoming integral to biomedical research and healthcare, revolutionizing multiple aspects of diagnosis, treatment, and patient care. Recent advances in the field, notably in generative AI (1,2), have significantly expanded AI capabilities, enabling personalized, precise, predictive, and portable healthcare delivery (3) including in low-resource settings where millions have inadequate access to essential healthcare services (4). Moreover, the potential economic impact of AI in healthcare is substantial, with estimates suggesting significant effects on healthcare spending (5). In many applications, particularly, medical image diagnosis, AI has outperformed human experts (3).

Despite these advancements, AI can exhibit and perpetuate inherent social biases amplifying health inequities (6,7). This is especially concerning in healthcare, where biased AI decisions can directly impact patient outcomes and exacerbate existing health disparities. The stakes are high because, unlike other domains, errors in healthcare AI can have life-or-death consequences. Bias in computer systems refers to systems that "systematically and unfairly discriminate[s] against certain individuals or groups of individuals in favor of others" (8). Bias may originate from various sources and model development stages. Data biases, algorithmic biases, and user interaction biases are often the main sources of AI bias (9,10). In healthcare, these biases can manifest in the form of minority bias, label bias, and clinician and patient biases resulting from AI interactions (11). Importantly, the interconnection between the sources of bias can exacerbate the impact of bias within AI systems, leading to a compounding effect that makes the bias more pervasive and harder to detect and mitigate (10).

Various solutions and initiatives are being undertaken to address bias and enhance fairness in AI. Algorithmic anti-discrimination laws and AI regulation are becoming a priority across states, countries, and regions. For instance, the US Department of Health and Human Services introduced the first federal civil rights law which mandates that covered entities do not discriminate against any individuals based on race, color, national origin, sex, age, or disability (12). However, in healthcare, "proving algorithmic bias ... is insufficient unless an actual injury to a plaintiff had resulted", and when an AI algorithm performs differently for different populations, it is challenging to "prove a physician should have known the output wasn't reliable for a particular patient" (13). Alongside legal and regulatory measures, ethical frameworks have been developed to guide ethical conduct in AI development (7,14–19), though some have criticized these frameworks for primarily reflecting Western cultural values and social norms (17,20). Technical solutions including bias auditing toolkits such as the IBM AI Fairness 360 Toolkit have been developed (21).

While technical solutions have been successful, particularly in healthcare (22), they are inherently limited (23) and challenging to apply in clinical settings (24). Critics argue that bias and fairness cannot be automated (25,26) and they call for behavioral changes in the healthcare industry to increase engagement in addressing these issues (27). This shift toward the "participatory turn" of AI (28) ensures that AI systems are not only designed by developers but also reflect the lived experiences and perspectives of end-users (29). Specifically, value-sensitive AI (VSAI) (30) aims to embody stakeholders' values in technology development. However, participation should not only seek data representativeness and completeness but also engage and empower those impacted by AI systems (24,28,31).

Prior works on AI bias and fairness (11,32,9,33,34,10) have focused on the types of bias, sources, and mitigation strategies. We complement these works by summarizing the limitations of current mitigation solutions, emphasizing healthcare and biomedical applications. These limitations include competing who defines bias and fairness – "who"; deciding which bias and fairness measures to use among dozens that are inconsistent and incompatible – "which", when in the model development are the measures most effective – "when", and for which populations – "to whom" and contexts – "where", they are designed. We illustrate each limitation with empirical studies and present solutions to address them. Moreover, we discuss how the VSAI framework developed for general-purpose AI can be adapted in the context of bias and fairness and how community engagement best practices in healthcare and biomedical research can inform this process.

1. **Successful, but Limited: From Social Concepts to Algorithms**

Technical bias mitigation solutions have successfully addressed racial bias in various clinical settings, such as in diagnosing burns from healthy skin, predicting mortality in intensive care units, forecasting healthcare expenditures, diagnosing diabetic retinopathy, classifying dementia, and predicting postpartum depression (22). However, empirical studies have revealed that despite substantial progress in AI research, translating these achievements into clinical settings remains challenging, with many AI investments stalling at the prototype level. For instance, studies have shown that approximately 22% of AI implementations demonstrated a direct impact on health outcomes, with the majority remaining in prototype testing phases. Among the remaining 78%, while AI models outperformed standard clinical modalities, they only indirectly influenced patient outcomes (35).

Technical bias mitigation solutions are often inspired by moral theories of justice and social sciences, such as Egalitarianism, Rawls' Principle of Equality, and Rawls' Minimax Principle (36–38). These techniques typically aim to achieve parity, maintain fairness in treatment or outcomes (38), or both (39). However, translating these principles into algorithms raises various ethical and technical challenges.

To understand the limitations, consider a simplified AI development pipeline consisting of design, implementation, and deployment phases. In the context of bias and fairness, the design phase involves making key algorithmic choices, such as defining what constitutes bias and fairness. After defining these concepts, the implementation phase seeks to translate them into algorithms and evaluate their effectiveness. This stage also connects to the AI training datasets, which are tied to the populations from which the data are collected. The deployment phase considers the real-world context in which these solutions are applied.

Thus, the conceptualization and formalization of bias and fairness, the effectiveness of the developed solutions, the AI training datasets, and the deployment environment all influence the success of technical approaches to bias mitigation. In the following, we discuss each limitation, extend them to the AI development stages in **Table 1**, and illustrate them with empirical studies, and solutions to address them.

### 1.1. Who Defines Bias and Fairness?

AI developers often dominate the design, implementation, and deployment of bias and fairness solutions. Critics argue that algorithms may reflect the political and ethical perspectives of their creators (40), carrying the creators' needs, values, and interests. This is concerning as the developers' interpretation of bias and fairness may conflict with those affected by the systems. Perceptions of fairness are influenced by factors such as algorithmic design, gender, age, education, AI literacy, AI transparency, and whether decisions are made by humans or machines (41). For example, seniority among future AI developers can lead to varied fairness interpretations (42). In healthcare, this issue is further complicated by medical interventions and patient outcomes (43). Further, while patients may prefer AI systems that do not consider protected characteristics like race or gender (41,44), clinicians might find such attributes clinically relevant and necessary for accurate diagnoses (43). Similarly, theoretical fairness may diverge from practical needs when existing fairness metrics do not align with individual preferences (45,46).

AI developers may also prioritize the performance and generalizability of the systems they develop (47,48) over ethical values such as fairness and overlook the potential negative impacts of the systems (49,50). This misalignment between developer priorities and healthcare needs creates a critical paradox: while developers may prioritize model performance metrics, healthcare requires models that are both accurate and equitable across diverse patient populations. This tension is exacerbated by the fact that healthcare decisions often involve multiple stakeholders – patients, providers, insurers, and healthcare systems – each with potentially conflicting fairness definitions. For example, a model optimized purely for accuracy might recommend more expensive treatments that insurers resist covering, while patients and providers might prioritize treatment effectiveness regardless of cost. This is concerning when unbiased AI systems can improve performance (51), and overlooking bias may hinder this potential.

To mitigate this issue, a growing literature suggests diversity in AI developers (52) as the lack of diversity in the AI workforce can amplify AI-generated inequalities (53). While efforts are being made toward this goal, the AI workforce still lacks diversity. For instance, in 2021, Hispanic, Black, and African Americans account only for 3.2% and 2.4% of new AI PhDs in the US (53). Diversity can address the misalignment of interests between developers and end-users, particularly when perceptions vary based on cultural, social, and individual experiences. Diversity should not be restrained to the AI workforce and should extend to include the voices of those impacted by AI systems. These voices can provide feedback or participate in the solution development process, offering valuable perspectives on how the systems affect their lives and ensuring that the fairness criteria align with their lived experiences (25).

Various frameworks have been developed for value trade-offs among stakeholders. For instance, in the context of fairness, the Stakeholder's Agreement on Fairness framework (25) can facilitate these trade-offs through an iterative process emphasizing ongoing stakeholder agreements. The method incorporates various perspectives to challenge power imbalances and biases throughout the AI system's development. By continuously engaging stakeholders, AI developers can ensure that fairness is not viewed as a one-time achievement but rather as an ongoing negotiation that adapts to new challenges and insights. This process includes identifying, mitigating, and monitoring biases at each stage of development, allowing developers to balance competing fairness goals. Importantly, these trade-offs are explained transparently to users, acknowledging that perfect fairness is unattainable, but the aim is to improve fairness iteratively. From a clinician-patient standpoint, (54) developed a value-based framework to guide clinicians in incorporating patients' values, notably equity, access, and justice, in using AI in clinical care.

Alongside the "who" dimension, the choice and prioritization of bias and fairness measures are also significant limitations of technical solutions, which will be addressed in the following section.

### 1.2. Which Bias Mitigation Strategies to Use and Prioritize?

The conflicting views on bias and fairness have led to the development of dozens of technical mitigation strategies (21,55,56). This raises concerns about which mitigation strategy to use and prioritize and whether the measures should be statistical or causal (57). Statistical metrics aim to mitigate unfairness by minimizing correlations between protected characteristics, non-protected characteristics, and outcome variables in AI training data. Causal techniques, like counterfactual fairness (58), ask whether changing an individual's protected characteristic would alter the outcome. This approach can embody human-in-the-loop principles (59) and has recently been successfully implemented to quantify health outcome disparities in invasive methicillin-resistant staphylococcus aureus infection (60). The choice between statistical and causal approaches in healthcare is complex because health outcomes often involve complex causal relationships. For example, a statistical correlation between race and treatment outcomes might mask underlying socioeconomic factors, making causal approaches more appropriate for healthcare applications.

However, causal approaches require more detailed data and stronger assumptions about the relationships between variables, which may not always be available or valid in clinical settings. Therefore, choosing between statistical and causal approaches requires careful consideration of the clinical context and the availability of reliable data.

Varying bias and fairness formalization have also resulted in mathematically inconsistent and incompatible measures (57,61,62). Inconsistency occurs when the metrics produce conflicting results and incompatibility, often known as the Fairness Impossibility Theorem (61,62) when the metrics cannot be used simultaneously. These foundational properties raise concerns about which metrics to use at the implementation stage of AI development. When metrics produce conflicting results, several issues emerge: the decision-making becomes more complex, the trust in AI can be eroded, and prioritizing one metric over another can favor certain groups. Moreover, different stakeholders might view the system as unfair as discussed earlier, which can lead to skepticism and resistance. Conflicting metrics can also be problematic by affecting compliance with legal, regulatory, and ethical standards.

Strategies for choosing between bias mitigation solutions have also been explored. Makhoulf et al. defined eleven selection criteria including the availability of ground truth in AI training data, the cost of classification, and existing regulations and standards (63). In mental health applications, Sogancioglu et al. found that reweighing and data augmentation techniques are best suited respectively for violence risk assessment and depression phenotype recognition for their properties to maintain fairness-accuracy tradeoffs (43). Foster et al. in the Fairness Tree defined three axes of selection depending on the nature of interventions: punitive, assistive, and resource-constrained settings (64). For instance, the authors suggest using group-level recall in resource-constrained settings where healthcare systems suffer. Deciding between statistical or causal techniques can be addressed using statistical metrics when bias is evident in measurable patterns, such as imbalanced data distributions, unequal error rates, or disparities across demographic groups. If the bias stems from complex cause-effect relationships or hidden confounders that are not captured by simple statistical patterns, a causal approach is more appropriate. To address metrics compatibility, AI developers can use the Maximal Fairness theorem (65) to determine sets of compatible metrics. However, these techniques remain techno-centric and should account for stakeholders' perspectives in the selection process (64).

Once bias and fairness measures are selected, determining when they are most effective is another challenge, which will be discussed in the following section.

### 1.3. When Are the Bias Mitigation Strategies Most Effective?

The effectiveness of bias and fairness metrics depends on varying factors including the tasks, models, choice of protected characteristics, and types of metrics (66). There is an ongoing debate about when the measures are most effective in AI model development (52). Bias and fairness interventions can be categorized into pre-processing, in-processing, and post-processing (22). Pre-processing techniques address bias before model training by removing protected characteristics, resampling data, or adjusting labels. These methods aim to create fairer training datasets that reduce the likelihood of biased outcomes later. In-processing techniques allow for fairness adjustments to be built directly into the model's structure. They can be explicit or implicit (39). Explicit methods enhance fairness by adjusting the loss function or applying fairness-aware algorithms (39), this includes Counterfactual Fairness. Implicit unfairness mitigation refers to algorithms that detect and address bias by adjusting the data representations learned during the training process, mainly used in deep learning models. Examples of implicit techniques include Adversarial Debasing (67). Importantly, these techniques can be used when protected characteristics are unavailable in the training data (39). Post-processing techniques focus on adjusting model outputs after training or modifying decision thresholds (22). This approach is often used when it is difficult to modify the training process or when fairness needs to be improved in already deployed models.

While each stage offers valuable approaches, their effectiveness varies depending on the problem being addressed. Pre-processing is critical when data bias is the main issue, in-processing is ideal for models requiring fairness integration, and post-processing is useful for fine-tuning models already in use. However, balancing fairness across these stages can be complex, and selecting the right technique requires careful consideration of the specific AI application and ethical trade-offs (68). Empirical evidence from healthcare implementation studies suggests that the effectiveness of bias mitigation strategies varies significantly across development stages. Studies have shown that AI applications require careful planning and strategies that transform both care services and operations to realize benefits (69). Moreover, assessing AI's value proposition must extend beyond technical performance and cost considerations to include an analysis of real-world care contexts (70).

In healthcare settings, the timing of bias mitigation is critical due to the dynamic nature of medical data and evolving clinical guidelines. Pre-processing approaches may need regular updates to reflect changing population demographics, while post-processing methods might require adjustment as clinical standards evolve. For instance, a model trained on historical data might need continuous pre-processing updates for changing disease patterns or treatment protocols. Additionally, the choice of timing must consider the operational constraints of healthcare systems, where model retraining might need to be balanced against the need for continuous availability of AI systems for clinical decision support.

Population descriptors highly influence the effectiveness of technical bias mitigation solutions and their formalization by AI developers is becoming a growing concern.

### 1.4. For Which Populations Are the Metrics Designed?

Bias and fairness measures often address fairness for individuals and groups. Individual-oriented techniques ensure that similar individuals receive similar outcomes. Group-oriented methods, on the other hand, ensure that similar outcomes are achieved across groups. Improving fairness for individuals might benefit certain subgroups but may not preserve fairness at the group level (71). However, group and individual fairness metrics can complement when properly defined and applied (72). Besides this apparent conflict, concerns arise about the availability of protected characteristics, their formalization, and intersectionality as discussed in the following.

#### 1.4.1. Availability And Formalization of Protected Characteristics

Group-oriented bias mitigation strategies require protected characteristics for the models to be "aware" of potential biases (71). However, in practice, these characteristics may be unavailable due to missing data or regulatory restrictions (73–76). For instance, most health insurers do not collect them (77). When such characteristics are available, there are often preferences against using them in AI development to ensure fairness (41). The healthcare context presents challenges for protected characteristics because certain demographic factors may have clinical relevance. This creates a tension between the need to consider these factors for accurate diagnosis and treatment, and the goal of preventing discriminatory outcomes (78). For example, genetic predispositions to certain conditions may correlate with racial categories (73) yet using race as a proxy for genetic variation risks perpetuating harmful stereotypes and may miss important individual variations. Healthcare AI systems must balance the clinical utility of demographic information against the risk of reinforcing biased care patterns.

In response, group-blind fairness techniques have been developed to improve fairness when protected characteristics are either unavailable or not explicitly used. These techniques often focus on capturing proxies of protected characteristics, for instance, by treating data regions with high and low AI errors as distinct groups (79) and using correlations between non-protected features and outcomes (75,80). Nevertheless, these techniques can still suffer from the proxy problem (81), as they may unintentionally allow non-protected features that are correlated with sensitive attributes to influence the model's decisions (82).

When protected characteristics are used, inconsistency in their formalization can be problematic (47). This is a growing concern in AI, with some arguing that "the algorithmic fairness community is an emerging race-making institution" (47). Others argued that using race and ethnicity as population descriptors is concerning as they are often social constructs (83) and can influence identity representation (84). For instance, categorizing individuals strictly by race or ethnicity can oversimplify complex identities and overlook important cultural nuances. To address these issues in genetics and genomic research, the National Academies of Sciences, Engineering, and Medicine developed a framework (85) for the use of population descriptors in biomedical research relevant to the context of algorithmic fairness. The framework emphasizes factors such as the data source, the scope of the data (individual-level and group-level), and the availability of pre-existing descriptors in the decision-making of population descriptors selection.

### 1.4.2. The Dilemma of Protected Characteristics Intersectionality

Traditional group fairness metrics typically address a single protected characteristic at a time. However, prioritizing fairness for one characteristic can lead to unfairness in others, highlighting a major limitation of group fairness metrics. For instance, increasing fairness for one characteristic may reduce fairness for others by up to 88.3% (86). This issue arises because individuals often possess multiple protected characteristics. Intersectionality is a problem not unique to algorithmic fairness but also identified in genomic research (87), and social sciences (88–90). The challenge for algorithmic fairness and AI design is determining which characteristics to prioritize and to what extent. In intensive care medicine, this selection should account for historical (e.g., racial), social disparities, and stakeholder dialogue (91). Various techniques have been developed to address intersectionality including metrics inspired by intersectionality frameworks from the Humanities (92,93). Other technical approaches include multi-calibration and multi-accuracy to address fairness across overlapping protected characteristics (94,95), and metrics that apply across different phases of AI development (96). We refer the readers to (97) for more details.

### 1.4.3. The Need for New Bias Mitigation Validation Datasets

The AI training, validation, and deployment datasets can affect the effectiveness of bias and fairness measures. AI developers typically rely on publicly and privately available datasets to validate the metrics. However, these datasets have notable limitations. Popular public datasets such as the Adult, German Credit Approval, and COMPAS datasets are noisy and include outdated racial categories (98). For example, the "Asian Pacific Islanders" racial category in the Adult dataset can obscure health disparities within these categories (99). These limitations highlight the need for more current bias and fairness training, validation, and deployment datasets. The All of US (100) program which aims to increase diversity in biomedical research datasets can be a candidate for such datasets. While private and commercial datasets might address some of the issues, they can also lead to problems with reproducibility and generalizability, especially in healthcare (101).

### 1.5. Which Context Are the Metrics Designed For?

AI systems should be closely connected to the real-world scenarios in which they will be deployed because they are highly sensitive to the characteristics of the training data (42). A significant concern arises when there is a mismatch between the deployment data and the training data, which can lead to reduced model performance. In healthcare, for instance, AI models in the US are often trained using data from only a few states, such as California, Massachusetts, and New York (102). This can limit their effectiveness in other regions and make bias mitigation harder to implement. Bias and fairness measures that work in one context might be inadequate in another (103). For example, in glaucoma prediction models, Ravindranath et al. found variation in fairness and accuracy in different contexts (104). Schrouff et al. have developed techniques for auditing distribution shifts in medical contexts (105). Domain Adaptation and Domain Generalization techniques can also help models adapt to the differences between training and deployment environments (106,107). These techniques are particularly useful when protected characteristics are available only during training or deployment, or when these characteristics differ between the two stages (108,109). For example, race may be used in the training data but gender in the deployment data (110).

Another issue with current bias mitigation strategies is that they often rely on US anti-discrimination laws, which may not be applicable in other cultural contexts (47). Effective bias mitigation requires culturally sensitive approaches (111) that account for local factors such as literacy, education, language, and the rural-urban divide, as seen in the African healthcare context (112,113). It is also essential to align justice and social science theories, which inspire bias and fairness solutions, with local moral frameworks. For example, in non-Western societies like Africa, moral traditions that prioritize community needs over individual ones may be more appropriate (114).

In summary, while technical bias mitigation strategies offer valuable tools, their limitations highlight the need for a more collective approach that includes stakeholder engagement and consideration of ethical, social, and contextual factors.

2. **Addressing AI Bias and Fairness Collectively**

There is increasing recognition that AI bias and fairness are not solely technical challenges, but socio-technical issues requiring collaborative approaches (24,27,31,77,115,116). Technologies must serve the values of end-users, a concept gaining traction across fields like technology design, political science, and social sciences (28). Several key frameworks and theories stress the importance of aligning technology with human values, focusing on usability, ethics, and the social contexts of users. Participatory Design, for example, as discussed by Schuler and Namioka (117), encourages involving end-users directly in the design process to ensure that the technologies reflect their values and social realities. Similarly, Value-Sensitive Design (VSD) promotes integrating moral and social values into technological design (118).

Participation is essential in ensuring that AI systems reflect the lived experiences and perspectives of end-users, rather than being the product of developers (29). This participatory approach is crucial for addressing AI bias and fairness issues (24,27,31,115,116). Human-centered AI (HCAI), in contrast to the biomedical community's engagement practices, advocates for broader stakeholder involvement in AI development. Chen et al. demonstrate how diverse healthcare stakeholders can collaborate to address AI bias at various stages of model development (116). Freedman et al. show how HCAI can help align a kidney exchange algorithm with human values (119).

Similarly, value-sensitive AI (VSAI), a subfield of HCAI, enhances technical solutions by integrating stakeholder values (30). The framework consists of three stages: conceptual, empirical, and technical (30). The conceptual stage involves identifying relevant stakeholders and understanding their context. The empirical and technical stages focus on gathering stakeholders' needs and values and embedding those values into the technology (here, technical bias solutions). This framework, developed for general-purpose AI can be adapted for addressing algorithmic bias collectively. We provide further recommendations later in the manuscript.

Various frameworks exist for identifying stakeholders in AI development at the conceptual stage. For instance, Miller's extension of the Stakeholder Salience Model uses four dimensions: power, legitimacy, urgency, and harm, to identify stakeholders' influence and relevance (120). The European Center for Not-for-Profit Law's framework for Meaningful Engagement offers similar strategies based on factors such as who is directly or indirectly impacted, who possesses subject matter expertise, and who has relevant experience (121). Overall, stakeholders can be categorized into three groups (122): individuals (users, engineers, researchers, non-users), organizations (technology companies), and national/international entities (lawmakers, regulators). Frameworks from community engagement in biomedical research, such as Community-Based Participatory Research (CBPR) (123), provide a basis for stakeholder identification and engagement in bias mitigation efforts. For example, in a recent study on AI-assisted cardiovascular screening, (124) employed the CBPR framework to engage clinicians and patients in the design process, resulting in a system that better addressed the needs and concerns of all stakeholders. Adus et al. developed a framework for patient engagement in healthcare AI development (125). The authors found that education is critical for meaningful patient engagement and patient preferences for in-person and social media-based recruitment channels.

At the empirical stage, diverse participatory methods such as Delphi interviews, citizens' juries, role-playing, and workshops can be used to collect stakeholder values (29). Schwartz's Basic Values framework can help guide these efforts by defining a set of universal human values that can be incorporated into the design (126). While the technical stage then ensures that these values are reflected in the technology, conflicts may arise, such as differing fairness values as discussed earlier. These can be resolved through the Stakeholders' Agreements on Fairness framework.

Despite its potential, participation in AI development has limitations that must be addressed for it to be effective (29). These include ensuring a diverse representation of stakeholders, managing conflicting interests, and integrating different perspectives into technical solutions. Participation should be purpose-driven and cannot solve all issues or replace democratic governance for public decisions (29). Moreover, inclusion does not always result in meaningful participation, as systemic barriers can limit marginalized groups' engagement. Cooptation is another risk, where grassroots efforts may be exploited by corporate interests, potentially disempowering communities. Measuring the effectiveness of participatory methods is also challenging, as many benefits are intangible and long-term. Additionally, power imbalances between stakeholders can distort genuine engagement. The US kidney allocation system is an example of such limitations, where, despite participatory efforts, bias persists in favor of certain patient groups (127).

These limitations underscore the need for careful and strategic approaches to ensure that participatory AI is genuinely inclusive and effective in addressing bias.

**Conclusion**
AI bias and fairness are critical concerns due to their potential to exacerbate and perpetuate health disparities. While technical solutions have been developed to tackle these issues, they face several limitations. This includes determining who defines bias and fairness, choosing and prioritizing appropriate metrics that may be inconsistent and incompatible, identifying the most effective timing for applying these metrics, specifying the target populations for which they apply, and considering the relevant context. Participatory AI provides a way to address these limitations by involving key stakeholders throughout different stages of the AI development lifecycle. We briefly discussed how the VSAI framework can be adapted to address bias collectively. However, this approach still faces challenges and remains theoretical. Ultimately, stakeholder engagement should not be seen as an alternative to technical solutions, but as a necessary complement that enriches the bias and fairness metrics and decision-making processes in AI systems. Future research should explore the validity of HCAI methods such as VSAI in real-world settings and the development of standardized frameworks that integrate technical and socio-ethical considerations. Additionally, investigating the role of emerging technologies like explainable AI and federated learning could provide new avenues for enhancing fairness in AI systems.

To address key questions about technical AI bias and fairness discussed here, the following recommendations can further be made for adapting the VSAI framework:

- **Who Defines Bias and Fairness?**
    - Identify key stakeholders for collaborative bias mitigation, for instance using Miller's extension of the Stakeholder Salience Model (120).
    - Establish a multidisciplinary working group from the identified stakeholders.
    - Use bias impact assessment frameworks to collaboratively evaluate the risk of bias in AI systems (for instance (128)).
    - Define stakeholders' fairness values for instance from Schwartz's universal values (126).
    - Agree on fairness objectives, for instance, using the Stakeholders' Agreement on Fairness framework (25)
    - Create ongoing channels for user feedback to refine definitions of fairness and bias continuously.
- **Which Bias Mitigation Strategy to Use and Prioritize?**
    - Engage stakeholders from different demographic, cultural, and professional backgrounds to collaboratively select bias and fairness metrics that align with diverse societal values and needs.
    - Collaboratively decide whether the solutions should be statistical or causal based on the application, technical constraints, and patients' and end-users' needs and values.
    - Ensure the selected metrics are consistent and compatible, such as through the use of the Maximal Fairness theorems (65).
    - Prioritize solutions that are flexible and adaptable to various contexts, ensuring that they reflect the specific needs of the end-users.
    - Regularly review and update fairness metrics for new ethical concerns and evolving social norms.
    - Evaluate the cost of the metrics to the AI models' accuracy and robustness to balance accuracy-fairness trade-offs.
    - Ensure that stakeholder's values are embodied in the metrics.
- **When Are the Bias Mitigation Strategies Most Effective?**
    - Evaluate bias and fairness metrics through pilot studies or simulations that test their performance in real-world scenarios before full implementation.
    - Use longitudinal studies to assess the long-term effectiveness of selected metrics, ensuring they remain relevant and effective over time.
    - Incorporate dynamic monitoring systems that assess the ongoing effectiveness of fairness metrics across different phases of AI system deployment.
    - Benchmark the bias mitigation solutions across different model development stages and select appropriate metrics.
- **For Which Populations Are the Metrics Designed?**
    - Define the target populations clearly, ensuring that fairness metrics are inclusive and account for the needs of underrepresented or marginalized groups.

- ○ Use participatory approaches to directly engage end-users in determining the solutions most applicable to their unique contexts and experiences.
  - ○ Ensure that the solutions account for intersectionality by considering how various identity factors (race, gender, socioeconomic status, etc.) interact.
  - ○ Ensure that the solutions training, validation, and deployment datasets, notably, the fairness attributes reflect current and context-specific protected population characteristics.
  - ○ Use the National Academies of Sciences, Engineering, and Medicine guidelines for the use of population descriptors (85) where appropriate.
- **Which Context Are the Metrics Designed For?**
  - ○ Tailor fairness metrics to the specific application areas (e.g., healthcare) to ensure they address context-specific ethical concerns and biases.
  - ○ Incorporate social, cultural, and historical considerations when designing metrics, ensuring relevance to the communities where AI systems are deployed.
  - ○ Use scenario planning and stakeholder input to anticipate future contexts in which fairness metrics might need adjustment or reevaluation.
  - ○ Ensure that the models' development and deployment are similar. Otherwise, clearly state the limitations of the bias mitigation strategies or use appropriate domain adaptation and generalization techniques.
  - ○ Ensure that those deploying the solutions understand both the technical complexities and the social implications of bias.

**DISCLOSURE STATEMENT**
The authors are not aware of any affiliations, memberships, funding, or financial holdings that might be perceived as affecting the objectivity of this review.

**ACKNOWLEDGMENTS**
Funding was provided by the GSK.ai-Stanford Ethics Fellowship (A.J.D.M and A.A.T.).

|  | Design | Implementation | Deployment |
|---|---|---|---|
| Who | Who defines bias and fairness? | Who implements and validates the solutions? | Who deploys solutions? |
| Which | Which bias mitigation strategy to use and prioritize?<br><br>Should the solutions be correlational or causal and who decides?<br><br>What is the cost of fairness to models' accuracy and robustness? | Are the metrics consistent and compatible? | Which metrics to deploy? |
| When | Are the metrics most effective before training the AI models (i.e., pre-processing fairness metrics)? | Are the metrics most effective during the training of the AI models (i.e., in-processing fairness metrics)? | Are the metrics most effective after training the AI models (i.e., post-processing fairness metrics)? |
| To whom | Should the metrics enhance fairness for individuals or groups?<br><br>How are protected characteristics formalized?<br><br>Is intersectionality taken into account by the metrics? | On which populations are the metrics validated? | On which populations are the metrics deployed? |
| Where | For which context are the metrics designed? | In which context are the metrics validated? | In which context are the metrics deployed? |

**Table 1: Summary of issues arising in the design, implementation, and deployment of technical bias fairness solutions.**